\pdfoutput=1

\documentclass[11pt,table,dvipsnames]{article}

\usepackage{acl}
\usepackage{times}
\usepackage{latexsym}

\usepackage[T1]{fontenc}

\usepackage[utf8]{inputenc}

\usepackage{microtype}

\usepackage{inconsolata}
\usepackage{microtype}
\usepackage{latexsym}
\usepackage{dsfont}
\usepackage{booktabs} 
\usepackage{subfigure}
\usepackage{CJKutf8}
\usepackage{graphicx}
\usepackage{bigstrut}
\usepackage{multirow}
\usepackage{amsmath}
\usepackage{multirow, makecell, caption}
\usepackage{floatrow}
\newfloatcommand{capbtabbox}{table}[][\FBwidth]
\usepackage{xcolor}
\usepackage[normalem]{ulem}
\usepackage{amssymb}
\usepackage{amsfonts}
\usepackage{multicol}
\usepackage{tipa}
\usepackage{amsmath}
\usepackage{bm}
\usepackage[ruled,linesnumbered]{algorithm2e}
\usepackage{tikz}
\usepackage{paralist}
\usepackage{IEEEtrantools}
\usepackage[switch]{lineno}
\usepackage{enumitem}
\usepackage{multirow, makecell, caption}
\usepackage{arydshln}
\usepackage{verbatim}
\usepackage[normalem]{ulem}
\usepackage{amssymb}
\usepackage{amsfonts}
\usepackage{multicol}
\usepackage{amssymb}
\usepackage{pifont}
\usepackage{csquotes}
\usepackage{xspace}
\usepackage{paralist}
\usepackage{mdwlist}
\usepackage{subfigure}
\usepackage{makecell}
\usepackage{array}
\usepackage{bm}
\usepackage{colortbl}
\usepackage{dsfont}
\usepackage{url}
\usepackage{booktabs} 
\usepackage{graphicx}
\usepackage{tabularx}
\usepackage{amsmath}
\usepackage{bbm}
\usepackage{pgfplots}
\usepackage{soul} 
\usepackage{listings}
\usepackage[tableposition=top]{caption}
\usepackage{color,xcolor} 

\newcommand{\ie}{\textit{i.e.}\xspace}
\newcommand{\eg}{\textit{e.g.}\xspace}

\newcommand{\xmark}{\ding{55}}
\newcommand{\cmark}{\ding{51}}

\newcommand{\method}{\textsc{EasyTool}\xspace}

\makeatletter
\def\adl@drawiv#1#2#3{%
        \hskip.5\tabcolsep
        \xleaders#3{#2.5\@tempdimb #1{1}#2.5\@tempdimb}%
                #2\z@ plus1fil minus1fil\relax
        \hskip.5\tabcolsep}
\newcommand{\cdashlinelr}[1]{%
  \noalign{\vskip\aboverulesep
           \global\let\@dashdrawstore\adl@draw
           \global\let\adl@draw\adl@drawiv}
  \cdashline{#1}
  \noalign{\global\let\adl@draw\@dashdrawstore
           \vskip\belowrulesep}}
\makeatother

\author{Siyu Yuan\textsuperscript{\rm $^1$}\thanks{~~The first two authors have equal contributions. This work was done when the first author was an intern at
Microsoft Research Asia.},
Kaitao Song\textsuperscript{\rm $^2$}\footnotemark[1] \footnotemark[2],\\ 
\bf Jiangjie Chen\textsuperscript{\rm $^1$},
Xu Tan\textsuperscript{\rm $^2$},
Yongliang Shen\textsuperscript{\rm $^3$},
Kan Ren\textsuperscript{\rm $^2$},
Dongsheng Li\textsuperscript{\rm $^2$},
Deqing Yang\textsuperscript{\rm $^1$}\thanks{~~Corresponding authors.}\\
\textsuperscript{\rm $^1$}Fudan University,
\textsuperscript{\rm $^2$}Microsoft Research Asia,
\textsuperscript{\rm $^3$}Zhejiang University\\
\texttt{syyuan21@m.fudan.edu.cn},
\texttt{\{kaitaosong, xuta, dongsli\}@microsoft.com}\\
\texttt{syl@zju.edu.cn},
\texttt{\{jjchen19,yangdeqing\}@fudan.edu.cn}}

\title{
\method: Enhancing LLM-based Agents with Concise Tool Instruction}

\begin{document}

\maketitle

\begin{abstract}
There has been a rising interest in utilizing tools in applications of autonomous agents based on large language models (LLMs) to address intricate real-world tasks.
To develop LLM-based agents, it usually requires LLMs to understand many tool functions from different tool documentations. However, these documentations could be diverse, redundant, or incomplete, which immensely affects the capability of LLMs in using tools.
To solve this, we introduce \method, a framework transforming diverse and lengthy tool documentation into a unified and concise tool instruction for easier tool usage.
\method purifies essential information from extensive tool documentation of different sources, and elaborates a unified interface (\ie, tool instruction) to offer standardized tool descriptions and functionalities for LLM-based agents.
Extensive experiments on multiple different tasks demonstrate that \method can significantly reduce token consumption and improve the performance of LLM-based agents on tool utilization in real-world scenarios.
Our code will be available at \url{https://github.com/microsoft/JARVIS/tree/main/easytool}.
\end{abstract}

\section{Introduction}
\label{sec:intro}
Large language models (LLMs)~\cite{openai2023gpt4, Hugo2023LLaMa, Hugo2023LLaMa2, geminiteam2023gemini} have recently ignited the spark of LLM-based autonomous agents~\cite{shen2023hugginggpt, gravitas2023auto}, which aim to interact with the real-world scenarios and address complex user requests. 
A rising trend in enhancing their effectiveness is to endow them with the capability of using external tools~\cite{Timo2023Toolformer, shen2023hugginggpt, qin2023toolllm}.
To bridge the gap between LLMs and tool usage, agents usually first analyze a user request, conduct planning or reasoning to decompose it into sub-tasks, and then select the most suitable tools for execution to obtain the final answer. 
Therefore, improving LLMs' capability to use tools precisely has been critical to developing an autonomous agent.

\begin{figure}[t]
    \centering
    \includegraphics[width=\linewidth]{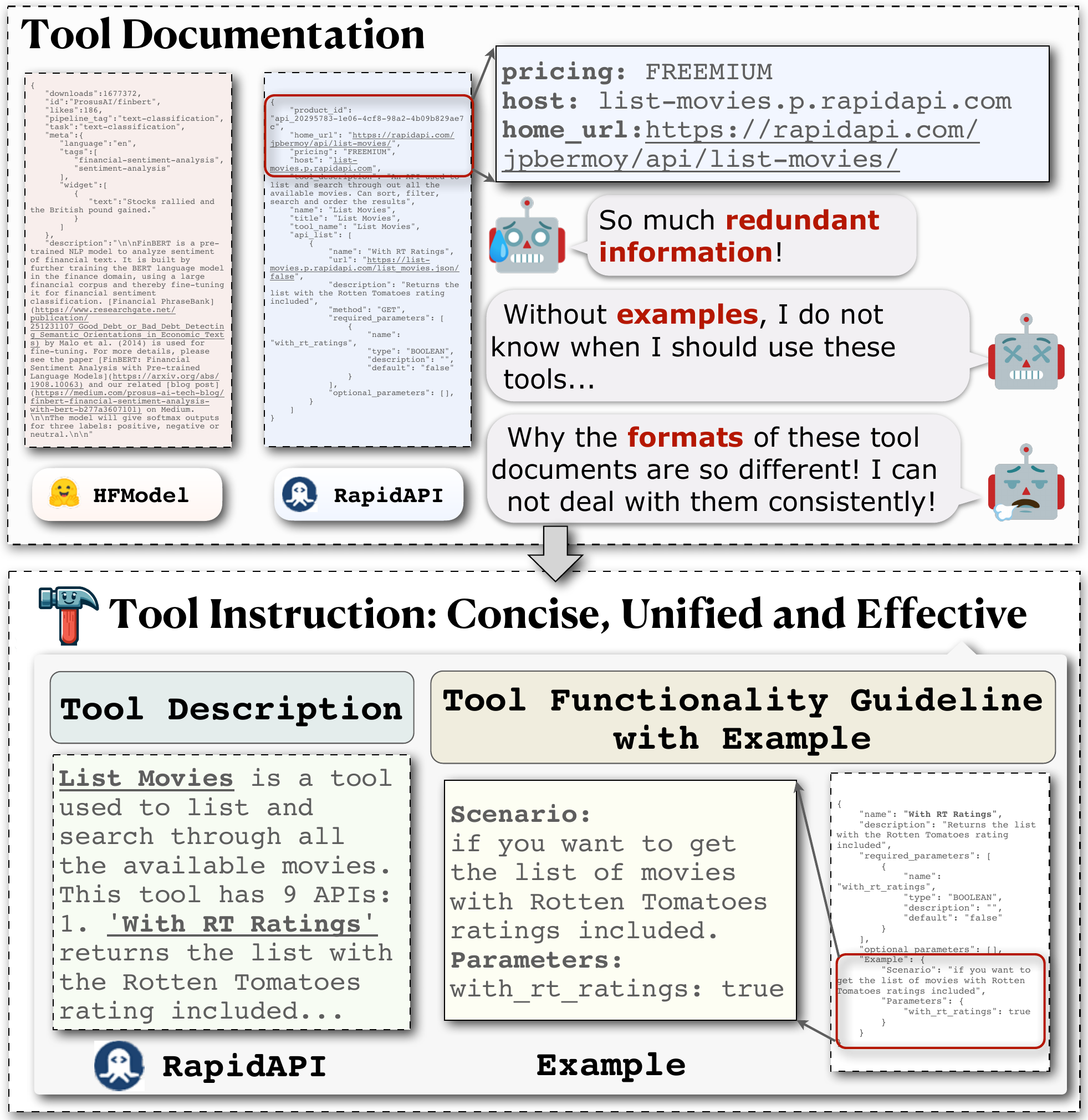}
    \caption{An illustration of the proposed \method, and some issues in tool documentation, \eg, Inconsistency, Redundancy, Incompleteness. The documentations can be polished and refined by \method into more concise and effective tool instructions for better tool usage.}
    \label{fig:front}
\end{figure}

Previously, some researchers~\cite{Timo2023Toolformer, qin2023toolllm, patil2023gorilla,parisi2022talm, hao2023toolkengpt} fine-tune open-source LLMs to generate calling functions to use tools. 
However, these methods usually require additional datasets with tool use for training, cannot be extended to widely deployed black-box LLMs (\eg, ChatGPT~\cite{openai2022chatgpt,openai2023gpt4} and Gemini~\cite{geminiteam2023gemini}), and lack flexibility in integrating external tools in a plug-and-play way. 
Another line of work~\cite{shen2023hugginggpt, song2023restgpt,lu2023chameleon, xu2023tool} retrieves and calls external tools by providing tool documentation and few-shot demonstrations of tool functionality. 
However, these methods struggle with limited context length and face difficulties when handling unusual tools, and thus hinder the development of an omnipotent LLM-based agent.
Therefore, extensive effort is still required to efficiently and effectively improve the quality of tool utilization.

For tool utilization, tool documentation plays an indispensable component, which could include multiple meta information like tool descriptions, tool parameters, demonstrations and so on.
However, as shown in Figure~\ref{fig:front}, we summarize the issues from existing documentation that could hinder the tool utilization of LLM-based agents:

\begin{itemize}[noitemsep, leftmargin=*]
    \item \textit{Inconsistency}:
    Massive tools from different sources often have inconsistent and diverse documentation formats, posing new challenges for LLMs to understand;
    \item \textit{Redundancy}:
    Tool documentation could encompass massive redundant and useless information, making it harder to grasp tool functionality and resulting in excessive token consumption in prompts;
    \item \textit{Incompleteness}:
    We expect the tool documentation to provide useful information to describe its functions, parameters and demonstrations for instructions. 
    However, the absence of critical information in some tool documentations impedes effective tool utilization.
\end{itemize}
Overall, we regard the information provided by tool documentation as a critical element in instructing LLMs to use tools.
However, the above issues in tool documentation bring some challenges to LLM-based agents to understand, especially considering the increasing of massive and diverse tools from different domains.
Therefore, how to parse the documentation, extract the most essential information and provide a unified format has become a necessary topic to effectively use tools.

In this paper, we introduce \method, an easy and effective method to create clear, structured, and unified instructions from tool documentations for improving LLM-based agents in using tools.
High-quality tool instructions should follow two criteria: easy to 1) understand its functionality for selection and 2) predict its parameters for usage.
To this end, we first collect massive tool documentations from different sources (\eg, RestBench~\cite{song2023restgpt} and ToolBench~\cite{qin2023toolllm}). 
Instead of directly using these various tool documentations with different complicated structures, we transform these documentations into a more concise and unified tool instruction, which includes standard \textit{tool descriptions} and guidelines for \textit{tool functionality}. 
The converted tool descriptions can eliminate irrelevant content and only keep the core functionality of each tool for LLMs to attend to.
Moreover, \method provides detailed information for tool usage (\eg, its parameters with demonstrations generated by ChatGPT~\cite{openai2022chatgpt}) in tool functionality guidelines to instruct LLMs with tool usage.

Extensive experiments on multiple datasets demonstrate these concise tool instructions generated by \method can significantly reduce incorrect tool usage.
Furthermore, we also prove that the capability of \method can be generalized to open-source LLMs in a plug-and-play way and greatly improve their performance on tool utilization in different real-world scenarios.
Our contributions can be summarized as:
\begin{itemize}[noitemsep]
    \item We analyze and explore the limitations of current tool utilization in LLM-based agents and first point out the deficiencies of tool documentation that hinder LLMs in using tools.
    \item To address these issues, we propose \method, which creates high-quality tool instructions from documentation to facilitate tool usage in LLM-based agents. 
    \item Experimental results on three datasets from distinct domains show that our \method effectively and efficiently improves the capability of LLMs in tool utilization.
\end{itemize}

\section{Related Work}
\label{sec:related}
With the emergence of powerful LLMs~\cite{openai2023gpt4, Hugo2023LLaMa, Hugo2023LLaMa2}, using tools has been considered a new trend to enhance the capabilities of LLMs
A conventional strategy is to build synthetic data~\cite{Timo2023Toolformer, qin2023toolllm, minghao2023APIBank, patil2023gorilla, shen2023taskbench} involved tool use and then fine-tune LLMs to generate text with tool invocation. 
However, these methods cannot be extended to some powerful closed LLMs, and lack the capability to use new tools. 
Although some methods~\cite{hao2023toolkengpt} attempted to fine-tune LLMs to obtain tool embeddings for plug-and-play usage, they still require additional data for training to get tool embeddings.

Therefore, there has arisen another branch~\cite{shen2023hugginggpt, song2023restgpt, gravitas2023auto, zhuang2024toolchain} that directly used LLMs as the controller and feed tool descriptions into prompts to instruct LLMs to understand and call tools. 
These methods do not need extra training and can use external tools in a plug-and-play paradigm, but they are limited to context sizes and the quality of tool documentation. 
As a result, these methods will lead to some failed or incorrect tool invocation~\cite{zhang2023syntax}. 
Some work~\cite{hsieh2023tool,xu2023tool} attempts to revise tool documentation to support a zero-shot tool utilization, but some inherent issues of tool documentation in real-world scenarios still hinder the effective and efficient usage of many tools.
Besides, different from naive prompt compression~\cite{mu2023learning,jiang-etal-2023-llmlingua}, which is only suitable to compress plain prompt, the streamlined information from tool documentation should satisfy specific format and need to confirm the accuracy of tool invocation when processing user requests.

\section{Preliminary}
\label{sec:challenge}
\begin{figure}[t]
    \centering
    \includegraphics[width=0.9\linewidth]{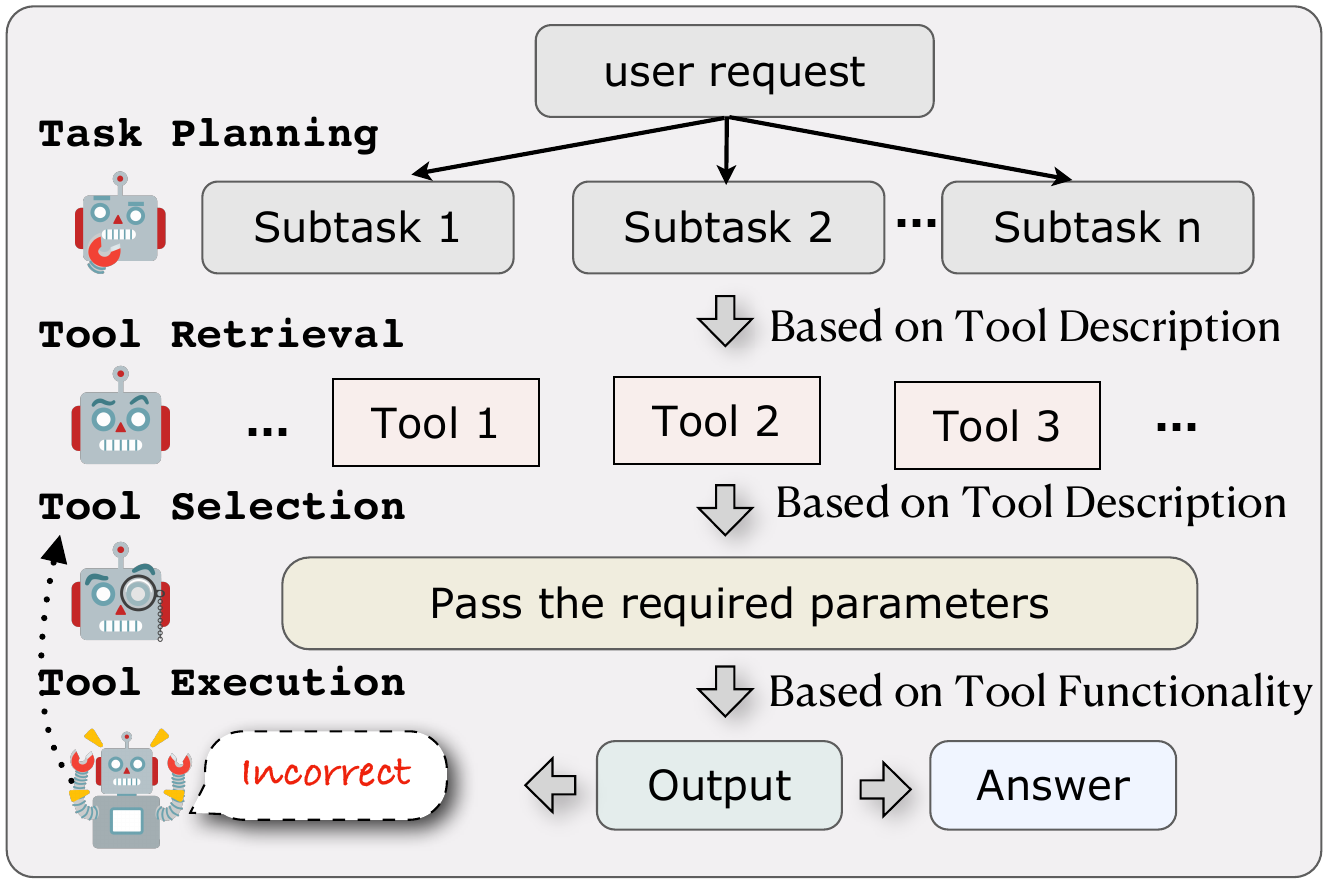}
    \caption{The four-stage framework of LLM-based agents in tool-usage applications.}
    \label{fig:tool_work}
\end{figure}

In this section, we first present the formulation to define the tool utilization in LLM-based agents and the limitations.

\subsection{Task Formulation}
Motivated by previous works~\cite{shen2023hugginggpt, song2023restgpt}, just as shown in Figure~\ref{fig:tool_work}, the pipeline of LLM-based agents for tool utilization can be summarized as a four-stage framework as:

\begin{itemize}[leftmargin=*,itemsep=2pt]
    \item \textbf{Task Planning:} Agents analyze a user request $\mathcal{T}$ and decompose it into subtasks $\mathcal{T}=\{t_1, t_2,..., t_n\}$ with specific dependencies and execution orders, each optimized for execution with a single tool.
    \item \textbf{Tool Retrieval:} Here, the focus is on matching these subtasks with suitable tools from the tool inventory based on the similarity between the subtasks and tools. The aim is to select the top-K tools $\{a_1, a_2,..., a_K\}$, that have the highest similarity to each subtask, forming a set of candidate tools for each.
    \item \textbf{Tool Selection:} In this stage, the most appropriate tool for each subtask from the set $\{a_1, a_2,..., a_K\}$ is chosen based on its description. This stage also includes preparing the parameters for tool execution, as specified in its document.
    \item \textbf{Tool Execution:} After tool selection and parameter setup, the tool is executed. If a tool fails during execution, the process reverts to the tool selection stage for an alternative choice. This retry mechanism continues until successful execution or until the maximum trial $\mathcal{R}$ is reached.
\end{itemize}
After these stages, agents can orchestrate different tools and use their powers to generate the final answer for each user request.

\begin{table}[t]
\small
  \centering
    \begin{tabular}{p{3.722cm}ccc}
    \toprule
    \textbf{Dataset} & \textbf{Token$_{\texttt{Desc.}}$} & \textbf{Token$_{\texttt{Doc.}}$} & \textbf{Exp.}\\
    \midrule
    RestBench~\cite{song2023restgpt} & 58 &3,881 & \xmark\\
    Gorilla~\cite{patil2023gorilla} & 88 &284 & \xmark\\
    ToolAlpaca~\cite{tang2023toolalpaca} & 567 & 7,661 & \xmark\\
    ToolBench~\cite{qin2023toolllm} & 744 &2,530 & \xmark\\
    HFmodels~\cite{shen2023hugginggpt} & 777 &1,196 & \cmark\\
    \bottomrule
    \end{tabular}%
  \caption{
  The statistics of tool documentations in tool benchmarks. 
  We report the average length of the tool description with parameters (\textbf{Token$_{\texttt{Desc.}}$}), the average length of the tool documentations (\textbf{Token$_{\texttt{Doc.}}$}) and whether the benchmarks have tool usage scenarios and example (\textbf{Exp.}).
  }
  \label{tab:redundancy}
\end{table}

\subsection{Analysis}\label{sec:limitation_analysis}
Previous studies typically adhere to the established paradigm, instructing LLMs with the tool documentation to use tools.
However, relying on the tool documentation can hinder the performance of LLM-based agents due to its inherent limitations.

\paragraph{Inconsistency}
In the real world, a wide variety of tools from different sources results in substantial diversity in terms of format, style, and guidelines.
As a result, this diversity contributes to a mess of tool documentations without a cohesive and standardized structure, posing a significant challenge for LLMs to effectively use these tools.

\paragraph{Redundancy}
Generally, the tool documents from different communities usually contain redundant information (\eg, URLs, IDs, etc.). 
In practical application, we just require LLMs to understand the core function of the tool and then decide whether to use and how to use this tool. 
As shown in Table~\ref{tab:redundancy}, we analyze multiple tool-based benchmarks and the results reveal a high proportion of redundant information in many tool documentations. 
For example, the average length of tool documentations used in ToolBench is approximately 2,530 tokens in Table~\ref{tab:redundancy}.\footnote{We adopt \texttt{cl100k\_base} encoding. The code is in \url{https://github.com/openai/tiktoken}.
} 
This useless information can severely hinder LLMs from retrieving and selecting tools, leading to an incorrect tool invocation.
Moreover, LLMs are constrained by a maximum context length, yet tool documentation is typically lengthy.
This excessive length can limit the range of tool options available for LLMs to consider, posing a challenge for efficient tool selection.

\paragraph{Incompleteness}
Previous work has demonstrated that LLMs may pass invalid parameters, leading to tool execution failure~\cite{song2023restgpt, qin2023toolllm, zhang2023syntax, shen2023taskbench}.
As shown in Table~\ref{tab:redundancy}, unlike human-oriented instruction manuals that provide usage scenarios and examples, existing tool documentation typically lacks such context, only offering example codes for tool invocation or results.
This leads to LLMs struggling to know when and how to refer to the examples to pass the correct parameters, resulting in invalid parameters.

\section{Method}
\label{sec:method}
As aforementioned, polishing, streamlining, and enhancing the tool documentation is important to improve tool utilization in LLM-based agents. 
In this paper, we introduce \method, a simple method to condense tool documentation into more concise and effective tool instructions.
The overall workflow is illustrated in Figure~\ref{fig:front}. 
Our framework comprises two stages: the first stage is to re-organize the original tool documentation by eliminating the irrelevant information and only keeping the function description of each tool ($\mathsection$~\ref{sec:step1}). 
Afterwards, for each tool, we further design a functional-guideline instruction for LLMs and enable LLMs to further refine the tool documentation by providing parameters of each tool and simultaneously the examples to instruct LLMs for usage ($\mathsection$~\ref{sec:step2}).

\begin{table}[t]
\footnotesize
  \centering
    \begin{tabularx}{\linewidth}{X}
    \toprule
    \rowcolor[gray]{0.95}\multicolumn{1}{c}{\textbf{I: Tool Description Generation}} \\
    \midrule
    \makecell[l]{
    \color{gray}{/* \textit{I: Task prompt} */}\\
    Your task is to create a concise and effective tool usage\\ 
    description based on the tool documentation. You should\\ensure 
    the description only contains the purposes of the\\tool without irrelevant information. Here is an example:\\
    \color{gray}{/* \textit{Examples} */} \\
    \texttt{\{Tool Documentation\}} \\
    Tool usage description:\\
    \texttt{\{Tool\_name\}} is a tool that can \texttt{\{General\_Purposes\}}.\\ 
    This tool has \texttt{\{Number\}} multiple built-in functions:\\
    1. \texttt{\{Function\_1\}} is to \texttt{\{Functionality\_of\_Function\_1\}}\\
    2. \texttt{\{Function\_2\}} is to ...\\
    \color{gray}{/* \textit{Auto generation of tool description} */} \\
    \texttt{\{Tool Documentation of `Aviation Weather Center'\}} \\
    Tool usage description:\\
    \color[rgb]{0,0.39,0}{\textit{`Aviation Weather Center' is a tool which can provide official}}\\
    \color[rgb]{0,0.39,0}{\textit{aviation weather data...}}}\\
    
    \midrule
    
    \rowcolor[gray]{0.95}\multicolumn{1}{c}{\textbf{II: Tool Function Guidelines Construction}} \\
    \midrule
    \makecell[l]{\color{gray}{/* \textit{Task prompt} */}\\
    Your task is to create the scenario that will use the tool. \\
    1. You are given a tool with its purpose and its parameters \\
    list. The scenario should adopt the parameters in the list.\\
    2. If the parameters are null, you\\should set: \texttt{\{"Scenario": XX, "Parameters":\{\}\}}.\\ Here is an example:\\
    \color{gray}{/* \textit{Examples} */} \\
    \texttt{\{Tool\_name\}} is a tool that can \texttt{\{General\_Purposes\}}.\\
    \texttt{\{Function\_i\}} is to \texttt{\{Functionality\_of\_Function\_i\}}\\
    \texttt{\{Parameter List of Function\_i\}}\\
    One scenario for \texttt{\{Function\_i\}} of \texttt{\{Tool\_name\}} is:\\
    \texttt{\{"Scenario": XX, "Parameters":\{XX:XX\}\}}\\
    \color{gray}{/* \textit{Auto-construction for Tool Function Guidelines} */} \\
     \makecell[l]{
     `Ebay' can get products from Ebay in a specific country. \\ 
     `Product Details' in `Ebay' can get the product details for a \\
     given product id and a specific country.\\ 
     \texttt{\{Parameter List of `Product Details'\}}\\
     One scenario for `Product Details' of `Ebay' is:\\
     \color[rgb]{0,0.39,0}\textit{\{"Scenario": "if you want to know the details of the product}\\
     \color[rgb]{0,0.39,0}\textit{with product ID 1954 in Germany from Ebay",}\\
     \color[rgb]{0,0.39,0}\textit{"Parameters":\{"product\_id": 1954, "country": "Germany"\}\}}.
     }
     }\\
    \bottomrule
    \end{tabularx}
  \caption{Examples of prompt for ChatGPT for tool description generation and tool function guidelines construction.
  {\color[rgb]{0,0.39,0}{\textit{Green texts}}} are generated by ChatGPT.
  }
  \label{tab:instruction_prompt}
\end{table}
\subsection{Tool Description Generation}\label{sec:step1}
As described above, tool documentation usually includes plenty of irrelevant information that makes it difficult to understand practical usage for LLMs. 
Moreover, some tools that enable multiple built-in functions for different scenarios are not always comprehensively described. 
For instance, Google Maps offers both distance calculations and coordinate provision, but its description might not cover all functionalities.
To address this, we expect to use LLMs to polish and streamline these tool documentations and decode them into more concise and effective tool descriptions. 
Here, just as shown in Table~\ref{tab:instruction_prompt} (I), we design an instruction and require LLMs (\ie, ChatGPT) to convert tool documentation to summarize its general purpose by following the designed instruction. We also add extra demonstrations into the instruction to enhance the instruction-following of LLMs in parsing tool documentation.

\subsection{Tool Functionality Guidelines Construction}\label{sec:step2}
The tool descriptions generated in the previous step aid LLMs in tool retrieval and selection.
However, we still need to predict the correct parameters of each tool for a successful execution. 
Previous work~\cite{qin2023toolllm, zhang2023syntax, xu2023tool} also confirms that many open-source LLMs are still inadequate in executing tools, resulting in parameter errors.
Therefore, we further polish our tool descriptions in the first stage to supplement the parameters in the tool instructions. 
Here, we design another instruction that requires LLMs to extract parameters from tool documentation and then organize it into a structured output, thus facilitating LLMs to invoke tools. As shown in Table~\ref{tab:instruction_prompt} (II), we use ChatGPT to create examples, including scenarios and parameter names with values to demonstrate how to input parameters for different scenarios and enhance LLMs to precisely use tools.
To verify the quality of generated examples for the tool functionality guidelines, we input the parameters to execute the tools to confirm the correct input of parameters and the accuracy of results.

\subsection{Evaluation}
To assess the quality of tool descriptions, we select 100 examples from ToolBench and employ three annotators to evaluate their accuracy.
The results confirm the high accuracy of all generated tool descriptions.
To assess the plausibility of the scenarios, we also sample 100 tool functionality guidelines from ToolBench and employ three annotators to evaluate the plausibility of the scenarios.
The results show that all of the scenarios are reasonable.
The annotation details for quality evaluation of tool instruction are shown in Appendix~\ref{sec:Crowd-sourcing}.\footnote{ 
The evaluation on the robustness of the prompts in Table~\ref{tab:instruction_prompt} is shown in Appendix~\ref{sec:robustness}.
We also compare with prompt compression methods in Appendix~\ref{appendix:compression}.
}

\section{Experiment}
\label{sec:exevalu}
In this section, we adopt \method to three distinct tool-use applications to show that \method can help LLM-based agents better answer real-world user requests through tool usage ($\mathsection$~\ref{sec:qa}), find correct tool solution paths ($\mathsection$~\ref{sec:movie}), and improve their tool utilization capabilities on complex math problems ($\mathsection$~\ref{sec:math}).

\begin{table}[t]
\small
  \centering
    \begin{tabular}{lccc}
    \toprule  
    \textbf{Dataset} & \textbf{Token$_{\texttt{Doc.}}$} & \textbf{Token$_{\texttt{Ins.}}$}  & \textbf{Reduce (\%)} \\
    \midrule
ToolBench & 2,530 & 748   & 70.43\% \\
RestBench & 3,881  & 103   & 97.35\% \\
\bottomrule
\end{tabular}
\caption{The average number of tokens in each tool documentation (\textbf{Token$_{\texttt{Doc.}}$}) and tool instruction generated by \method (\textbf{Token$_{\texttt{Ins.}}$}). We also report the reduced ratio (\ie \textbf{Reduce (\%)}) for reference.}
\label{tab:tool_doc}
\end{table}

\newcolumntype{a}{>{\columncolor{BlueGreen!10}}c}
\newcolumntype{b}{>{\columncolor{Green!10}}c}
\newcolumntype{d}{>{\columncolor{Gray!10}}c}
\newcolumntype{q}{>{\columncolor{Blue!10}}c}

\begin{table*}[t]
\small
  \centering
    \begin{tabular}{lcqqqbbbddd}
    \toprule
    \multirow{2}[0]{*}{\textbf{Model}} & \multirow{2}[0]{*}{\textbf{Method}} & \multicolumn{3}{c}{\textbf{I2-Category}} & \multicolumn{3}{c}{\textbf{I3-Instruction}} & \multicolumn{3}{c}{\textbf{Average}}\\
    \cmidrule(lr){3-5}
    \cmidrule(lr){6-8}
    \cmidrule(lr){9-11}
       &   & \multicolumn{1}{c}{\textbf{Pass}} & \multicolumn{1}{c}{\textbf{Win}} & \multicolumn{1}{c}{\textbf{Success}} & \multicolumn{1}{c}{\textbf{Pass}} & \multicolumn{1}{c}{\textbf{Win}} & \multicolumn{1}{c}{\textbf{Success}} & \multicolumn{1}{c}{\textbf{Pass}} & \multicolumn{1}{c}{\textbf{Win}} & \multicolumn{1}{c}{\textbf{Success}}\\
    \midrule
    \multirow{4}[0]{*}{ChatGPT} &ReACT & 39.0  &    -   & 18.0  & 23.0  &   -    & 1.0 & 31.0  & -   & 9.5  \\
     & DFSDT & 64.5  & 63.0  & 24.0  & 60.0  & 70.0  & 6.0  & 62.3  & 66.5  & 15.0  \\
     & DFSDT-\method & 74.5 & 76.5 & \underline{68.5} & 65.0 & 88.0 & 37.0 & 69.8  & \underline{82.3}  & 52.8  \\
     & DFSDT-\method-Retriever & 69.0  & 71.0  & 60.5  & {66.0}  & \underline{89.0}  & \underline{42.0}  & 67.5  & 80.0  & 51.3  \\

    \midrule
    \multirow{4}[0]{*}{ToolLLaMA-7B} & ReACT & 30.0  & 45.5  & 9.5  & 22.0  & 49.0  & 3.0 & 26.0  & 47.3  & 6.3  \\
     & DFSDT & 66.0  & 55.0  & 24.0  & 56.0  & 56.0  & 6.0 & 61.0  & 55.5  & 15.0 \\
     & DFSDT-Retriever & 57.0  & 60.0  & 11.5  & 54.0  & 69.0  & 2.0  & 55.5  & 64.5  & 6.8  \\
    \cdashlinelr{1-11}
    \multirow{4}[0]{*}{Vicuna-7B} & ReACT & 0.0  & 0.0  & 0.0  & 0.0  & 0.0  & 0.0  & 0.0   & 0.0   & 0.0  \\
     & DFSDT & 0.0  & 0.0  & 0.0  & 0.0  & 0.0  & 0.0   & 0.0   & 0.0   & 0.0  \\
     & DFSDT-\method & 72.5 & \underline{77.0} & 40.5 & \underline{68.0} & 81.0 & 34.0 & 70.3  & 79.0  & 37.3  \\
      & DFSDT-\method-Retriever & \underline{75.0}  & 68.0  & 46.5  & {67.0}  & 85.0 & 36.0  & 71.0  & 76.5  & 41.3  \\

    \cdashlinelr{1-11}
    \multirow{4}[0]{*}{Mistral-Instruct-7B} & ReACT & 0.0  & 0.0  & 0.0  & 0.0  & 0.0  & 0.0  & 0.0   & 0.0   & 0.0  \\
     & DFSDT & 0.0  & 0.0  & 0.0  & 0.0  & 0.0  & 0.0   & 0.0   & 0.0   & 0.0  \\
     & DFSDT-\method & \underline{75.0}  & 76.0  & 56.0  & 66.0  & 87.0  & 38.0  & 70.5  & 81.5  & 47.0   \\
      & DFSDT-\method-Retriever & 74.5  & 71.5  & 54.5  & \underline{68.0}  & 88.0  & 38.0  & \underline{71.3}  & 79.8  & 46.3  \\
    \midrule
    
    \multirow{4}[0]{*}{GPT-4} &ReACT & 67.5  & 53.5  & 27.0  & 40.0  & 71.0  & 4.0 & 53.8  & 62.3  & 15.5  \\
      &DFSDT & 69.5  & 57.0  & 42.0  & 59.0  & 73.0  & 50.0 & 64.3  & 65.0  & 46.0  \\
     &DFSDT-\method & \textbf{76.5} & \textbf{78.5} & \textbf{76.0} & \textbf{69.0} & \underline{89.0} & \textbf{64.0} & \textbf{72.8}  & \textbf{83.8}  & \textbf{70.0}  \\
     & DFSDT-\method-Retriever & 72.5 & 72.0 & 73.5 & \textbf{69.0} & \textbf{90.0} & 53.0 & 70.8  & 81.0  & \underline{63.3}  \\
    \bottomrule
    \end{tabular}%
  \caption{Results of LLMs on ToolBench. Win rate (denoted as \textbf{Win}) is calculated by comparing each model with ChatGPT-ReACT. 
  The win rate higher than 50\% means the model performs better than ChatGPT-ReACT.
  Apart from Retriever, all methods use the ground truth toolset to select tools.
  The best results are \textbf{bolded}, and the second best ones are \uline{underlined}. }
  \label{tab:toolbench}
\end{table*}%

\subsection{Real-World Question Answering}\label{sec:qa}
Since LLMs are still limited to their training data, it is essential for LLMs to use external tools to access up-to-date information in response to user requests.

\paragraph{Benchmark}
We choose ToolBench~\cite{qin2023toolllm}, a dataset containing diverse user requests with a massive set of publicly available REST APIs spanning 49 categories from RapidAPI Hub. 
We use the most difficult subsets of ToolBench to evaluate our method, \ie, I2-Category (200 test data) and I3-Instruction (100 test data), which contain complex user requests that need multiple tools from different categories to solve. 
On average, an I2-Category request needs 6.76 tools, and an I3-Instruction request needs 8.24 tools. 
Each data sample of ToolBench consists of a user request with a ground truth toolset, and thus models only need to select and execute the tools from the toolset to complete the user request.

For evaluation, ToolBench designs two evaluation metrics based on ChatGPT:
\begin{inparaenum}[(1)]
    \item \textbf{Pass Rate}, calculated by the proportion of instructions successfully completed within a limited budget;
    \item \textbf{Win Rate}, measured by asking a ChatGPT evaluator to select its preference for two solution paths.
\end{inparaenum}
Furthermore, we also measure \textbf{Success Rate}, which asks GPT-4 to check whether the responses can reasonably and accurately answer the user requests.\footnote{The prompt template for evaluating success rate is shown in Appendix~\ref{sec:success_rate_prompt}}

\paragraph{Baselines}
Following \citet{qin2023toolllm}, we select ChatGPT~\cite{openai2022chatgpt}, GPT-4~\cite{openai2023gpt4}, Vicuna-7B~\cite{vicuna2023}, 
and ToolLLaMA-7B as baselines and apply both ReACT~\cite{yao2023react} and DFSDT~\cite{qin2023toolllm} to them.
ToolLLaMA-7B is fine-tuned from a 7B LLaMA model~\cite{Hugo2023LLaMa} on ToolBench.
We do not adopt ToolLLaMA-7B on \method due to its poor instruction-following capability.
Furthermore, we also adopt Mistral-Instruct-7B~\cite{jiang2023mistral} for comparison, which is fine-tuned from Mistral-7B and exhibits great instruction-following capability.\footnote{Detailed information of baselines is shown in Appendix~\ref{appendix:baselines}}
For the baselines, we follow the settings in \citet{qin2023toolllm}, which provides tool documentation for them to use tools.

\paragraph{Main Result}
We simplify the tool documentation from ToolBench into concise tool instructions with \method.\footnote{The data examples of ToolBench are provided in Appendix~\ref{sec:example_toolbench}.}
Each tool instruction consists of a tool description and functionality guidelines.
As shown in Table~\ref{tab:tool_doc}, with \method, replacing tool documentation with our tool instruction can greatly reduce the token cost of each tool. 
Especially in ToolBench, the token cost was reduced by 70.43\%.

Furthermore, results in Table~\ref{tab:toolbench} show that:
\begin{inparaenum}[\it 1)]
    \item With the \method generated tool instructions, LLMs can achieve state-of-the-art performance.
    Notably, ChatGPT + DFSDT-\method even surpasses GPT-4 + DFSDT in success rate, indicating the superiority of tool Instructions over tool documentation in facilitating tool utilization for LLMs; 
    \item Vicuna and Mistral-Instruct-7B result in a failure when directly using tools. Based on previous experiences~\cite{shen2023taskbench}, we attribute this phenomenon to the lack of training in formatted data (\eg, function call).
    However, tool instructions generated by \method can help these models to better understand the usage of tools, even making them outperform the fine-tuned method, \ie, ToolLLaMA;
    \item Mistral-Instruct-7B outperforms Vicuna-7B with \method, indicating that models with better instruction-following capabilities can achieve greater improvements with high-quality tool instructions.
\end{inparaenum}

\begin{table}[t]
\small
  \centering
    \begin{tabular}{lcccccc}
    \toprule
\multirow{2}[0]{*}{\textbf{Method}} & \multicolumn{2}{c}{\textbf{I2-Category}} & \multicolumn{2}{c}{\textbf{I3-Instruction}} & \multicolumn{2}{c}{\textbf{Average}} \\ 
\cmidrule(lr){2-3}
\cmidrule(lr){4-5}
\cmidrule(lr){6-7}
  & \textbf{@1} & \textbf{@5} & \textbf{@1} & \textbf{@5} & \textbf{@1} & \textbf{@5} \\ 
\midrule
BERT Retriever  & \underline{68.2} & \underline{77.9} & \textbf{81.7} & \underline{87.1} & \underline{75.0} & \underline{82.5} \\ 
GPT Ada & 36.8 & 30.7 & 54.6 & 46.8 & 45.7 & 38.8 \\
\ \ + \method & \textbf{73.4} & \textbf{82.7} & \underline{80.1} & \textbf{88.5} & \textbf{76.7} & \textbf{85.6} \\
\bottomrule
\end{tabular}
\caption{The performance of different retrievers for two subsets in ToolBench. We report \textbf{NDCG@1} and \textbf{NDCG@5}.}
\label{tab:retrieval}
\end{table}

\begin{figure}[t]
    \centering
    \includegraphics[width=\linewidth]{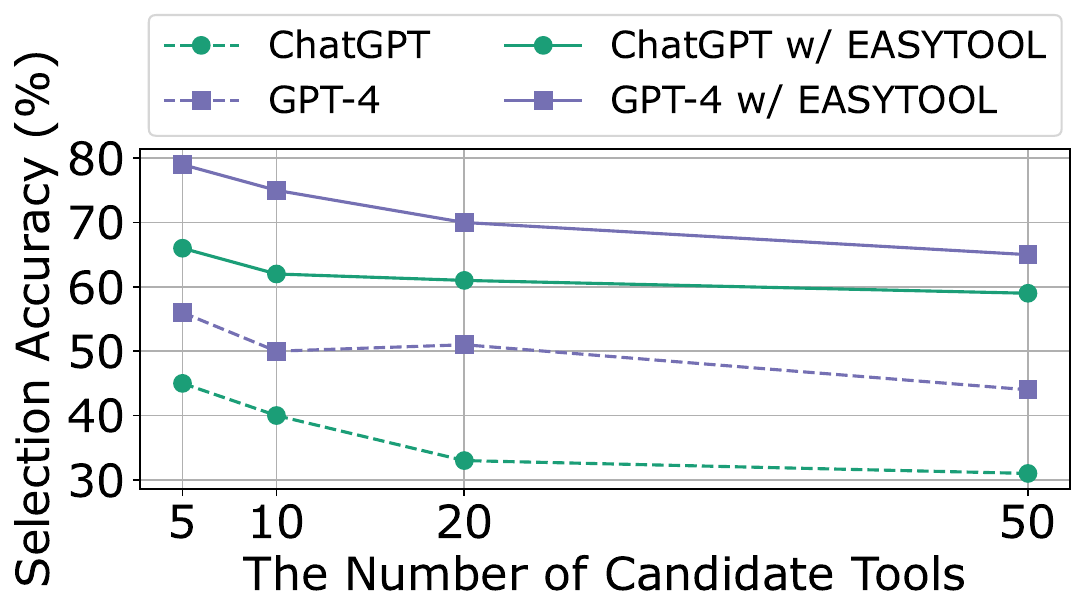}
    \caption{The selection accuracy of LLMs on I1-instruciton of ToolBench.}
    \label{fig:selection_accuarcy}
\end{figure}

\paragraph{\method helps retrieve high-quality tools.}
In real-world scenarios, asking users to manually recommend tools from a large pool for LLMs to select may not be practical.
Therefore, ToolBench also provides a dense retriever based on BERT-base~\cite{devlin-etal-2019-bert} to retrieve relevant tools for solving user requests, and claims that it outperforms \texttt{text-embedding-ada-002}, \ie GPT Ada~\cite{ouyang2022training}, which retrieves tools based on the cosine embedding similarity between the subtasks decomposed by user requests and tool descriptions from original tool documentation in ToolBench.
We argue that the poor performance of Ada may be due to low-quality tool descriptions, which often contain irrelevant details and lack clear functionality guidelines.
Thus, we adopt the tool description generated by \method to replace the original tool descriptions.
Following \citet{qin2023toolllm}, we compare the performance of these retrieval methods using NDCG~\cite{10.1145/582415.582418}.
Results in Table~\ref{tab:retrieval} show that providing tool descriptions generated by \method can greatly improve the retrieval performance.

\paragraph{\method helps LLMs with tool selection and execution.}
To answer this question, we utilize the I1-Instruciton of ToolBench, which comprises 100 user requests solvable by a single tool.
We first obtain the golden tool from I1-Instruction and then retrieve other different tools based on cosine embedding similarity between the user request and tool descriptions as candidate tools.
Then, we evaluate the selection accuracy of LLMs with varying numbers of candidate tools, using either original ToolBench descriptions or those generated by \method. 
Figure~\ref{fig:selection_accuarcy} illustrates that \method-enhanced descriptions enable LLMs to select the correct tool more effectively from a larger pool.

For each subtask in I2-Category and I3-Instruction, we retrieve the top 10 most similar tools using our tool descriptions and ask models to select and execute them.
As shown in Table~\ref{tab:toolbench}, using these retrieved tools proves to be comparable, and sometimes even superior, to the ground truth tool set. 
The rationale is that \method-Retriever can retrieve similar tools with better functionalities to replace some tools in the ground truth tool set.

\begin{figure}[t]
    \centering
    \includegraphics[width=\linewidth]{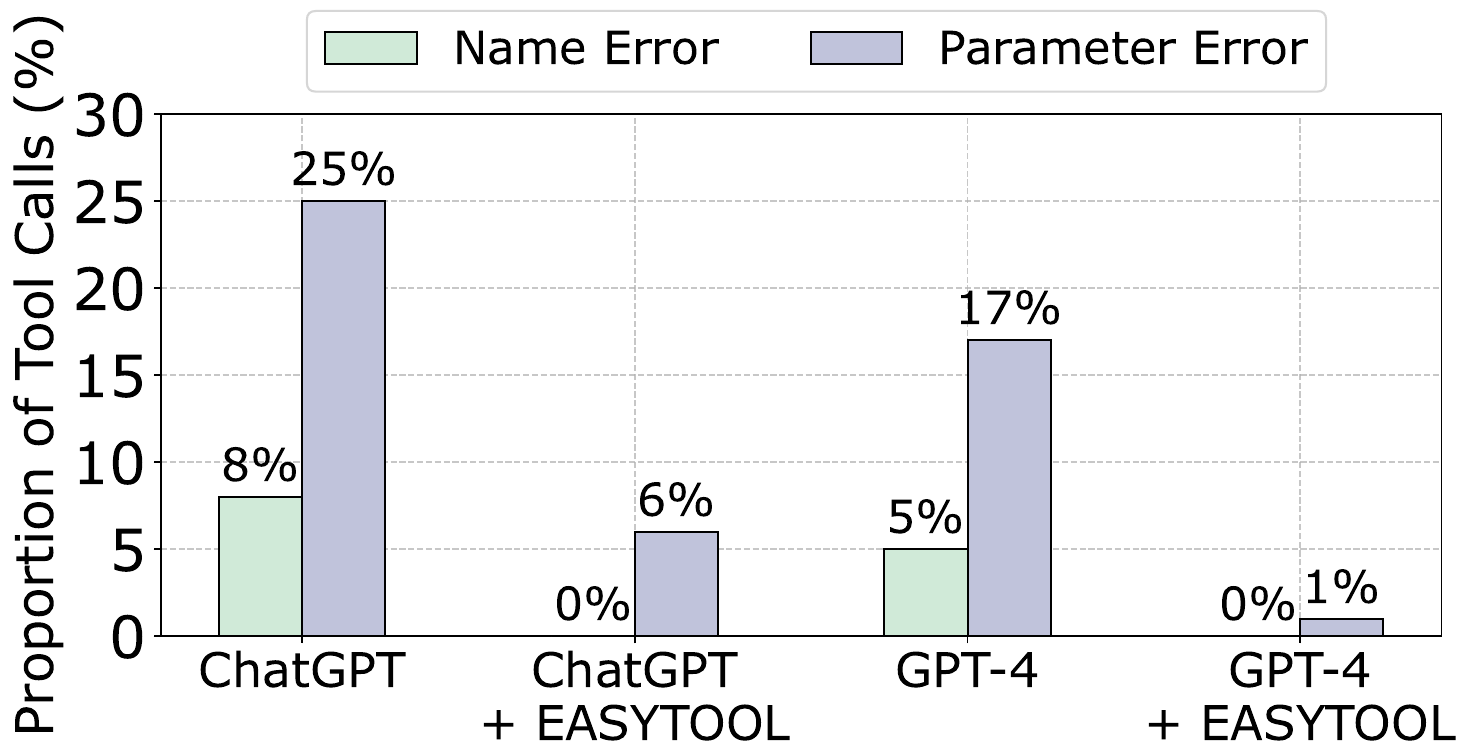}
    \caption{
    Error rates of tool calls in different LLMs.
    The error rate is the number of two tool-related errors relative to the total number of tool calls.
    The results are evaluated by \textbf{human annotators}.}
    \label{fig:error_analysis}
\end{figure}

\paragraph{Error Analysis}
We follow \citet{zhang2023syntax} and define two types of error, \ie, Tool name error and parameter error. 
Tool name error means models call non-existent tool functions that are not in the tool inventory, and parameter error means models pass invalid parameters. Both errors lead to unsuccessful tool execution.
We sample 100 data from I2-Category and I3-Instruction and employ three annotators to manually examine the output of LLMs with tool documentation and tool instruction generated by \method.
We present the error rates of each error type on ToolBench in Figure~\ref{fig:error_analysis}.
The results show that LLMs may generate non-existent tool names and pass invalid parameters to the right tool functions.
However, our \method, with concise and effective tool instruction, can significantly reduce these incorrect behaviors, leading to successful tool execution.

\subsection{Real-World Web Services}\label{sec:movie}
Real-world web services often need to execute tools following a specific order.
For example, a shopping cart Web Service requires users to add items to the cart before checkout.
We aim to explore the capability of LLMs to find correct tool solution paths.

\paragraph{Benchmark}
We select RestBench~\cite{song2023restgpt}, which consists of tasks in real-world web service scenarios.
We evaluate our method on a subset of RestBench, \ie, TMDB.
TMDB is a movie information website that offers 55 official RESTful APIs as tools, which cover information on movies, TVs, actors, and images.
Following the evaluation metric in RestBench, we use the correct path rate (CP\%) to measure accuracy, which is the proportion of the model-generated tool path containing the gold tool path as a subsequence.

\paragraph{Baselines}
We choose RestGPT~\cite{song2023restgpt} as our base model.
The RestGPT has two versions, \ie, Vicuna-13B-based RestGPT and ChatGPT-based RestGPT.
For Vicuna-13B-based RestGPT, we compare our method with ToolDec.
For ChatGPT-based RestGPT, we compare our method with ReAct since ToolDec cannot apply in close-sourced models.

\begin{figure}[t]
    \centering    
    \includegraphics[width=\linewidth]{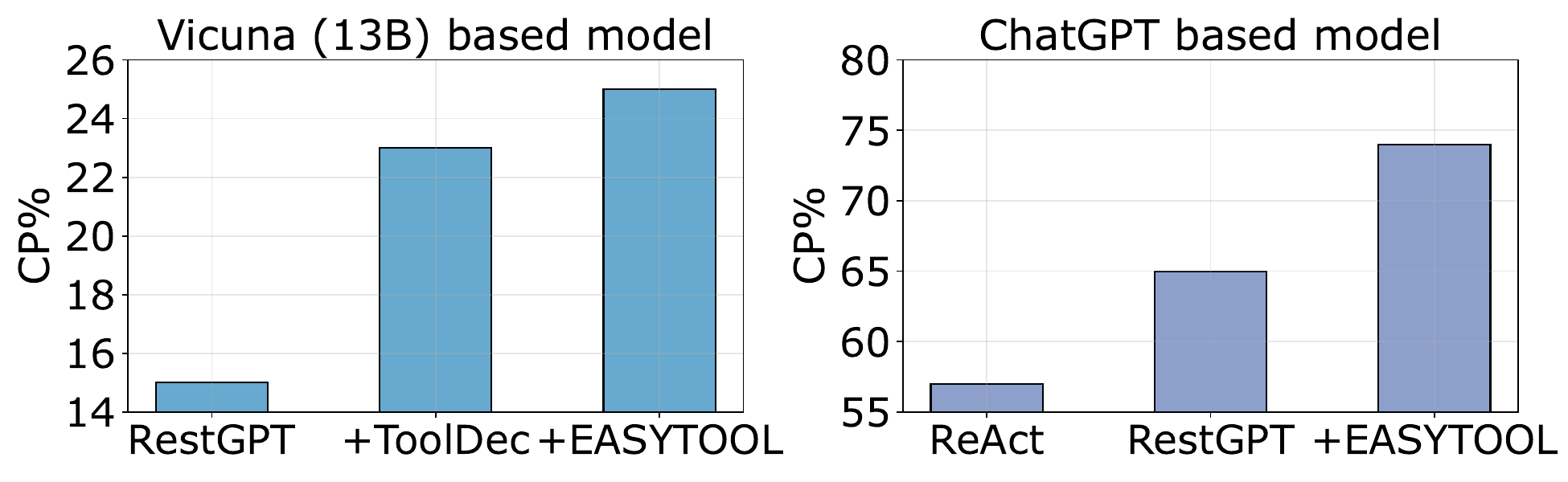}
    \caption{The correct path rate (CP\%) on two versions of RestBench with different methods.}
    \label{fig:restbench}
\end{figure}
\paragraph{Result}
We simplify the long tool documentation from RestBench into concise tool instructions with \method for LLMs to use.\footnote{The data examples of RestBench are provided in Appendix~\ref{sec:example_restbench}.}
For comparison, we use the prompt from \citet{song2023restgpt} containing original tool descriptions and four examples.
Table~\ref{tab:tool_doc} demonstrates that \method significantly reduces the token cost. 
Additionally, Figure~\ref{fig:restbench} highlights the considerable improvement in the correct path rate, signifying \method's effectiveness in aiding LLMs to find the correct tool solution paths.

\subsection{Numerical Reasoning}\label{sec:math}
We also explore whether \method can endow LLM-based agents with better tool-utilization ability in complex math problems with incomplete tool documentation.

\paragraph{Benchmark}
We adopt FuncQA~\cite{hao2023toolkengpt}, which tests the numerical reasoning ability of LLMs on complex math problems involving 13 arithmetic operations tools (\eg, multiply, power and lcm).
We use the two subsets of FuncQA. \ie \textbf{one-hop} and \textbf{multi-hop} questions, to evaluate our method.
The one-hop questions consist of 68 math problems solvable with one tool.
The 60 multi-hop questions require some reasoning steps, averaging 2.78 tool uses per question.
We measure accuracy by calculating the percentage of correctly answered problems, with a 0.1\% error tolerance. 
We also measure tool error rate (\ie, \textbf{Error}), the proportion of tasks that have at least one tool-related error.

\paragraph{Baselines}
Following \citet{hao2023toolkengpt}, We select Vicuna-30B and ChatGPT and compare our method with 0-shot learning, Chain-of-thought (CoT) prompting and ReAct. 
For ReAct, we follow the settings in \cite{hao2023toolkengpt}, which provide four examples, including five tool examples.

\begin{table}[t]
\small
  \centering
    \begin{tabular}{lccc}
    \toprule
    \textbf{Model} & \textbf{One-hop} ($\uparrow$)& \textbf{Multi-hop} ($\uparrow$)& \textbf{Error} ($\downarrow$)\\
    \midrule
    Vicuna-30B & 15.00 & 1.00 & -\\
    \ \  + CoT & 13.33 & 4.00 & -\\
    \ \  + ReAct & 45.00 & 7.35 & 20.31\\
    \ \  + \method & \textbf{65.00} & \textbf{11.76} & \textbf{10.15}\\
    \midrule
    ChatGPT & 55.00 & 9.00 & - \\
    \ \  + CoT & 48.33 & 17.64 & -\\
    \ \  + ReAct & 85.00 & 41.17 & 9.38\\
    \ \  + \method & \textbf{91.66} & \textbf{48.53} & \textbf{2.34}\\
    \bottomrule
    \end{tabular}%
  \caption{The accuracy of Vicuna-30B and ChatGPT on the FuncQA. 
  }
  \label{tab:funcqa}
\end{table}

\paragraph{Result}

Unlike the other datasets, FuncQA only provides the name and calling function of a tool as documentation, without any other tool descriptions for further usage demonstration. 
Therefore, by only leveraging the provided tool name and calling function, we can also apply \method to generate tool descriptions with usage scenarios to construct tool instruction for FuncQA.\footnote{
The data examples of FuncQA are provided in Appendix~\ref{sec:example_funcqa}.}
Results in Table~\ref{tab:funcqa} show that:
\begin{inparaenum}[\it 1)]
    \item The tool instructions generated based on our method (+ \method) significantly improve the tool utilization ability of LLMs on complex math problems;
    \item Furthermore, a lower tool error rate of models with \method indicates that compared to few-shot learning with demonstrations, concise and effective tool instructions can better guide models to select correct tools and pass valid parameters.
\end{inparaenum}

\section{Conclusion}
\label{sec:conclusion}

In this paper, we introduce \method, an easy and effective method to enhance the tool utilization capabilities of LLM-based agents through the simplification and refinement of tool documentation into a clear, structured and practical tool instruction.
Our comprehensive experiments demonstrate that \method can effectively enhance performance in different real-world applications.
We hope \method can be a significant development in the field of LLM-based agents.

\section*{Limitations}
\label{sec:limitation}
First, this paper only focuses on tool documentation whose token length does not exceed the ChatGPT input limit. 
Documentation with token counts surpassing this limit cannot be processed by \method without additional preprocessing.
Second, our method is limited to single documentation, neglecting the dependencies among tools. 
Considering these dependencies in tool descriptions could significantly enhance the model's effectiveness in certain scenarios.
Finally, \method only works on models with instruction-following ability. 
Future work can focus on training specialized models using tool instructions generated by \method, thereby improving their capability in tool utilization.

\section*{Ethics Statement}
\label{sec:Ethics}
We acknowledge that all authors are informed about and adhere to the ACL Code of Ethics and the Code of Conduct.

\paragraph{Use of Human Annotations}
Our institution recruited annotators to implement the annotations of tool descriptions and functionality guidelines. 
We ensure the privacy rights of the annotators are respected during the annotation process.
The annotators receive compensation exceeding the local minimum wage and have consented to the use of tool instructions generated by \method for research purposes. Appendix~\ref{sec:Crowd-sourcing} provides further details on the annotations.

\paragraph{Risks}
The tool benchmarks in our experiment are sourced from publicly available sources. However, we cannot guarantee that they are devoid of socially harmful or toxic language. 
Furthermore, evaluating the data quality of tool instructions is based on common sense, which can vary among individuals from diverse backgrounds.
We use ChatGPT to correct grammatical errors in this paper.

\bibliography{anthology,custom}

\clearpage
\appendix
\section{Crowd-sourcing Details for Tool Instruction Evaluation}\label{sec:Crowd-sourcing}
\begin{figure*}[t]
    \centering
    \includegraphics[width=\linewidth]{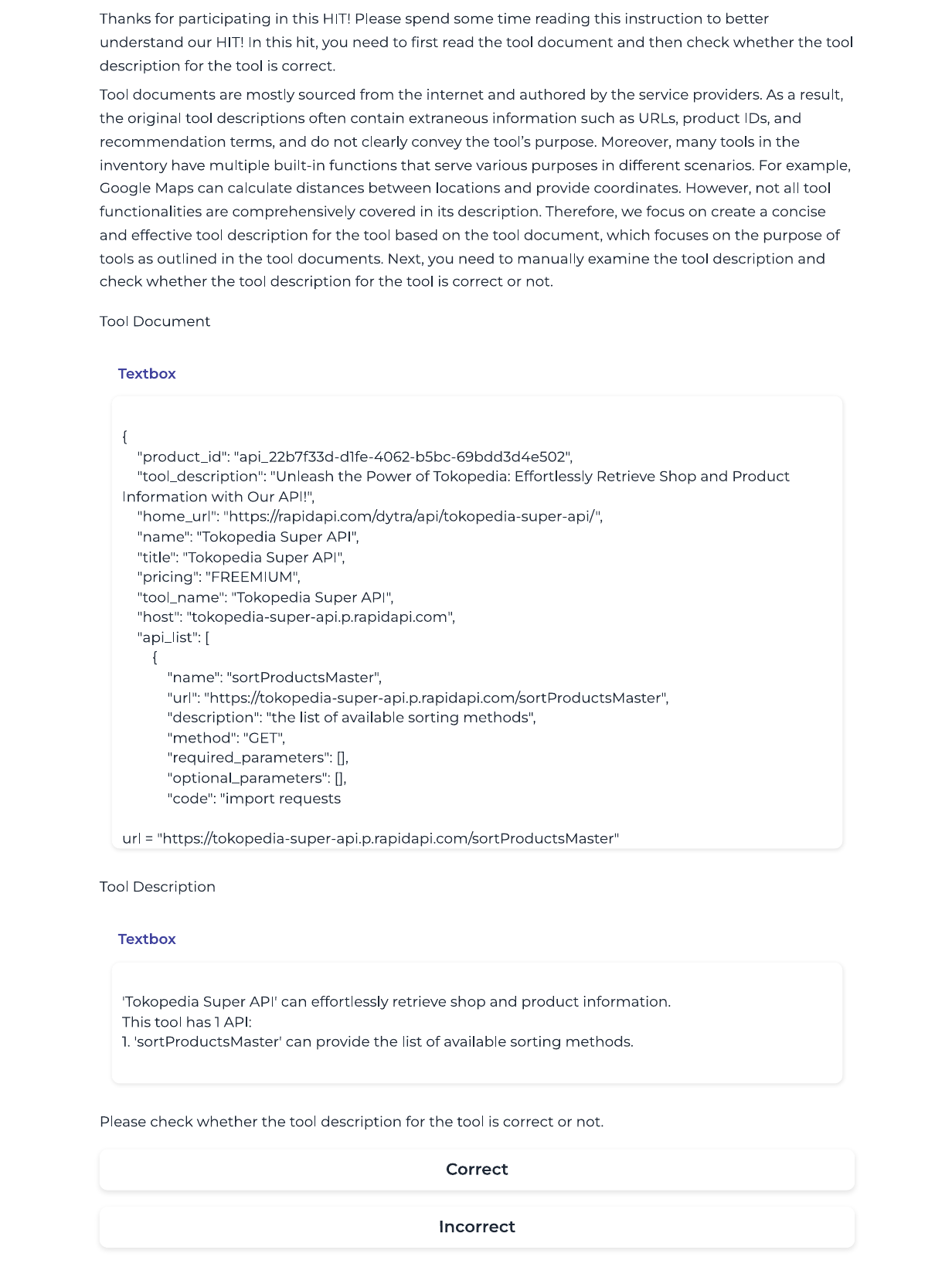}
    \caption{The screenshots of the instructions and interface for tool description annotation.}
    \label{fig:Gradio_description}
\end{figure*}

\begin{figure*}[t]
    \centering
    \includegraphics[width=\linewidth]{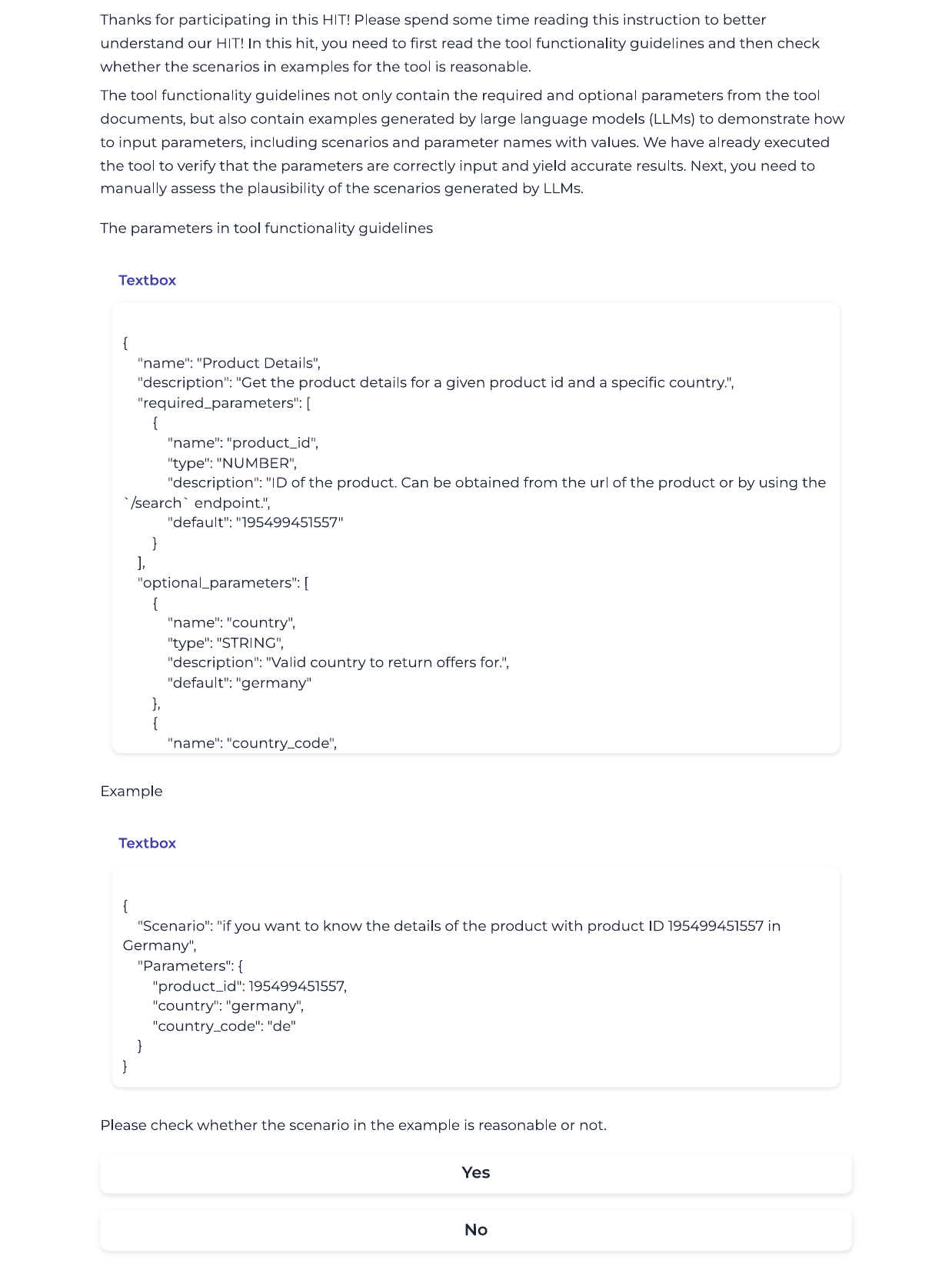}
    \caption{The screenshots of the instructions and interface for tool functionality guidelines annotation.}
    \label{fig:Gradio_parameter}
\end{figure*}

We have recruited a team of three undergraduates.
We pay each annotator \$8/h, exceeding the local minimum wage. 
The screenshots of the instructions and interface for tool description and functionality guideline annotation are shown in Figure~\ref{fig:Gradio_description} and Figure~\ref{fig:Gradio_parameter}.

\section{Details of ToolBench}

\subsection{Success Rate Evaluation}\label{sec:success_rate_prompt}
The prompt of the success rate evaluation is given in List~\ref{lst:success_rate_prompt}.

\lstset{
    backgroundcolor=\color[RGB]{245,245,244},
    breaklines=true,
    breakindent=0pt,
    basicstyle=\ttfamily\small,
    emphstyle={\bfseries\color{NavyBlue}}
}\begin{lstlisting}[caption={Instruction templates for GPT-4 to evaluate the success rate of the results on  ToolBench},label=lst:success_rate_prompt]
Please check whether the response can reasonably and accurately answer the question. If it can, please output 'YES'; If not, please output 'NO'

You need to give reasons first and then decide whether the response can reasonably and accurately answer the question. You must only output in a parsible JSON format. Two example outputs look like:

Example 1: {"Reason": "The reason why you think the response can reasonably and accurately answer the question", "Choice": "Yes"}
"Example 2: {"Reason": "The reason why you think the response cannot reasonably and accurately answer the question", "Choice": "No"}

This is the user's question: {question}
This is the response: {answer}
Output: 
\end{lstlisting}

\subsection{The Details of Baselines on ToolBench}\label{appendix:baselines}
Vicuna-7B~\cite{vicuna2023} is the LLaMA variant fine-tuned on instructions and user-shared conversations.
Mistral-Instruct-7B~\cite{jiang2023mistral} is the Mistral-7B variant fine-tuned on instructions, which exhibits great instruction-following and reasoning capability.
The ReAct in \citet{qin2023toolllm} first decomposes the use request into subtasks and then plans the tool calls to complete the subtasks.
The DFSDT in \citet{qin2023toolllm} adopts a depth-first search-based decision tree to enable LLMs to make deliberate decisions by assessing different reasoning paths.

Following the setting in \citet{qin2023toolllm},
for ChatGPT and GPT-4, we directly leverage the function call to use tools~\footnote{\url{https://openai.com/blog/function-calling-and-other-api-updates}}.
For other models, we synthesize input in function call format to these models.

\section{Robustness Evaluation}\label{sec:robustness}
\begin{figure}[t]
    \centering
    \includegraphics[width=0.8\linewidth]{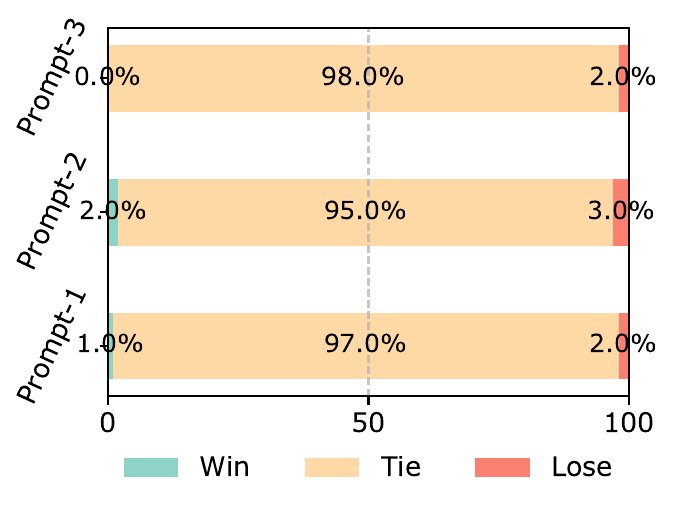}
    \caption{Comparison of our task prompts with ChatGPT generated task prompts. Percentage of wins, ties and losses are calculated.}
    \label{fig:win_rate}
\end{figure}

In this section, we aim to evaluate the robustness of the task prompts in Table~\ref{tab:instruction_prompt}.
We ask ChatGPT to rewrite these task prompts three times, and the new task prompts are shown in List~\ref{lst:robustness}.
We sample 100 tool documentations from ToolBench and ask ChatGPT to generate tool descriptions and tool functionality guidelines based on the new task prompts.
Then, we ask two annotators to evaluate the quality of four results (one from our task prompts and three from ChatGPT-generated task prompts). 
The results in Figure~\ref{fig:win_rate} show that the changes to the task prompt, without altering the actual meaning, do not affect the quality of the tool description and tool functionality guidelines, thereby demonstrating the robustness of our prompts.

\lstset{
    backgroundcolor=\color[RGB]{245,245,244},
    breaklines=true,
    breakindent=0pt,
    basicstyle=\ttfamily\small,
    emph={Prompt,-,1,2,3},
    emphstyle={\bfseries\color{NavyBlue}}
}\begin{lstlisting}[caption={The ChatGPT generated task prompts for tool description generation and tool function guidelines construction.},label=lst:robustness]
Prompt-1:
- Tool Description: 
Your assignment involves developing a succinct and practical description on using a specific tool, as outlined in its documentation. This description should focus solely on the tool's functions, excluding any extraneous details.
- Tool Function Guidelines:
Create a scenario that incorporates the use of a specified tool, ensuring it utilizes the provided parameters.
Receive a description of a tool, including its purpose and a list of parameters. Design a scenario that effectively employs these parameters.
If both "required_parameters" and "optional_parameters" are absent, format your response as:
{"Scenario": XX, "Parameters": {}}.

Prompt-2:
- Tool Description: 
Your assignment involves crafting a succinct and practical description of a tool, using its documentation as a reference. Focus on outlining the tool's functions, excluding any extraneous details.
- Tool Function Guidelines:
Your assignment involves developing a scenario that utilizes a specified tool. Here are the guidelines:
You will receive information about a tool, including its intended use and a list of parameters. Your scenario should incorporate these parameters.
In cases where both "required_parameters" and "optional_parameters" are absent, format your response as follows:
{"Scenario": XX, "Parameters": {}}.

Prompt-3:
- Tool Description: 
Your assignment involves writing a succinct and clear description of a tool's usage, guided by its documentation. This description should exclusively focus on the tool's functions, omitting any extraneous details.
- Tool Function Guidelines:
Your assignment involves crafting a scenario that utilizes a specific tool. Here's how to proceed:
First, familiarize yourself with the tool's intended use and its available parameters. Then, design a scenario that effectively incorporates these parameters.
In cases where both "required_parameters" and "optional_parameters" are absent, format your response as follows:
{"Scenario": XX, "Parameters": {}}.
\end{lstlisting}

\section{Prompt Compression Method}\label{appendix:compression}
We also adopt LLMLingua~\cite{jiang-etal-2023-llmlingua}, a prompt compression method, to identify and remove non-essential tokens in tool documentation.
As shown in Table~\ref{tab:LLMLingua}, this method can not be applied to our task since it may compress some tokens in parameters and functions, which are essential for successful tool execution.

\section{Examples of Tool Instruction}
\subsection{Data Examples of ToolBench}\label{sec:example_toolbench}
Table~\ref{tab:example_toolbench} presents some examples of tool instructions generated by \method in ToolBench for a better understanding.
\subsection{Data Examples of RestBench}\label{sec:example_restbench}
Table~\ref{tab:example_restbench} presents some examples of tool instructions generated by \method in RestBench for a better understanding.
\subsection{Data Examples of FuncQA}\label{sec:example_funcqa}
Table~\ref{tab:example_funcqa} presents some examples of tool instructions generated by \method in FuncQA for a better understanding.

\lstset{
    backgroundcolor=\color[RGB]{250,250,250},
    breaklines=true,
    breakindent=0pt,
    basicstyle=\ttfamily\small,
    emph={Tool,Documentation,Instruction,Compressed,By,LLMLingua},
    emphstyle={\bfseries\color{NavyBlue}},
}
\begin{table*}[!ht]
\begin{lstlisting}
Tool Documentation:
{
    "product_id": "api_b04d269d-c7dd-4b84-8e17-6fba24d64d3d",
    "tool_description": "Get Products from Ebay (Unofficial)",
    "home_url": "https://rapidapi.com/felixeschmittfes/api/ebay32/",
    "name": "Ebay",
    "title": "Ebay",
    "pricing": "FREEMIUM",
    "tool_name": "Ebay",
    "host": "ebay32.p.rapidapi.com",
    "api_list": [
        {
            "name": "Product Details",
            "url": "https://ebay32.p.rapidapi.com/product/195499451557",
            "description": "Get the product details for a given product id and a specific country.\nDefault country is `United States`.\nSpecify country with country name or country code.\n\nAllowed countries:\nDefault: `us`\n- Germany (de)\n- France (fr)\n- Australia (au)\n- Austria (at)\n- Canada (ca)\n- Hong Kong (hk)\n- Ireland (ie)\n- Italy (it)\n- Malaysia (my)\n- Netherlands (nl)\n- Singapore (sg)\n- Switzerland (ch)\n- United Kingdom (uk)",
            "method": "GET",
            "required_parameters": [
                {
                    "name": "product_id",
                    "type": "NUMBER",
                    "description": "ID of the product. Can be obtained from the url of the product or by using the `/search` endpoint.",
                    "default": "195499451557"
                }
            ],
            "optional_parameters": [
                {
                    "name": "country",
                    "type": "STRING",
                    "description": "Valid country to return offers for.\nValid values are in description of this endpoint.\nDefault: `united states`.",
                    "default": "germany"
                },
                {
                    "name": "country_code",
                    "type": "STRING",
                    "description": "Country code of the valid country to return offers for.\nValid values are in description of this endpoint.\nDefault: `us`.",
                    "default": "de"
                }
            ]
        }
    ]
}

Tool Instruction Compressed By LLMLingua:
{
  "product "_b04d269d-c7be-fba24d64d",
 "_ "Get fromay (Unofficial "://id./fixeschmittfes/ay/ " " " " "FREEM " " ".p. "_ [   " "Product Details",
            "url": "https://ebay32.p.rapidapi.com/product/195499451557",
            "description": "Get the product details for a given product id and a specific country.
Default country is `United States`.
Specify country with country name or country code.
\end{lstlisting}
\caption{The original tool documentation and tool instruction compressed by LLMLingua.}
\label{tab:LLMLingua}
\end{table*}

\lstset{
    backgroundcolor=\color[RGB]{250,250,250},
    breaklines=true,
    breakindent=0pt,
    emph={Tool,Description,Function,Guidelines},
    emphstyle={\bfseries\color{NavyBlue}},
    basicstyle=\ttfamily\small,
}
\begin{table*}[!ht]
\begin{lstlisting}
Tool Description:
/* Example 1 */
'TokopediaApi' can search and retrieve product details from Tokopedia. This tool has 2 APIs: 1. 'Search Product' can search for products on Tokopedia based on a query string and action type. 2. 'Get Product Detail' can retrieve detailed information about a product on Tokopedia based on its slug.
/* Example 2 */
'Tokopedia Super API' can effortlessly retrieve shop and product information. This tool has 1 API: 1. 'sortProductsMaster' can provide the list of available sorting methods.

Tool Function Guidelines:
/* Example 1 */
{
    "name": "Search Product",
    "description": "Search The Product",
    "required_parameters": [
        {
            "name": "query",
            "type": "STRING",
            "description": "",
            "default": "Celana Jeans"
        },
        {
            "name": "act",
            "type": "STRING",
            "description": "",
            "default": "search"
        }
    ],
    "optional_parameters": [],
    "Example": {
        "Scenario": "if you want to search for a product with the query 'Celana Jeans' using the 'search' action",
        "Parameters": {
            "query": "Celana Jeans",
            "act": "search"
        }
    }
}
/* Example 2 */
{
    "name": "sortProductsMaster",
    "description": "the list of available sorting methods",
    "required_parameters": [],
    "optional_parameters": [],
    "Example": {
        "Scenario": "if you want to retrieve the list of available sorting methods for products using Tokopedia Super API",
        "Parameters": {}
    }
}
\end{lstlisting}
\caption{The tool instruction of ToolBench generated by \method.
  }
\label{tab:example_toolbench}
\end{table*}

\lstset{
    backgroundcolor=\color[RGB]{250,250,250},
    breaklines=true,
    breakindent=0pt,
    emph={Tool,Description,Function,Guidelines},
    emphstyle={\bfseries\color{NavyBlue}},
    basicstyle=\ttfamily\small,
}
\begin{table*}[!ht]
\begin{lstlisting}
Tool Description:
/* Example 1 */
'/tv/latest' can get the most newly created TV show.
/* Example 2 */
'/search/collection' can search for collections, which can obtain collection_id.

Tool Function Guidelines:
/* Example 1 */
{
    "tool_usage": "GET /person/{person_id}/tv_credits",
    "Example": {
        "Scenario": "If you want to get the TV show credits of a person with person_id 456.",
        "Parameters": {
            "input": "GET /person/456/tv_credits"
        }
    }
}
/* Example 2 */
{
    "tool_usage": "GET /tv/latest",
    "Example": {
        "Scenario": "If you want to get the most newly created TV show.",
        "Parameters": {
            "input": "GET /tv/latest"
        }
    }
}
\end{lstlisting}
\caption{The tool instruction of RestBench generated by \method.
  }
  \label{tab:example_restbench}
\end{table*}

\lstset{
    backgroundcolor=\color[RGB]{250,250,250},
    breaklines=true,
    breakindent=0pt,
    emph={Tool,Description,Function,Guidelines},
    emphstyle={\bfseries\color{NavyBlue}},
    basicstyle=\ttfamily\small,
}
\begin{table*}[!ht]
\begin{lstlisting}
Tool Description:
/* Example 1 */
'add_' returns the sum of all the arguments passed to it, normalized to 2 decimal places.
/* Example 2 */
'subtract_' returns the difference of the arguments passed to it, starting with the first argument and subtracting all subsequent arguments, normalized to 2 decimal places.

Tool Function Guidelines:
/* Example 1 */
{
    "required_parameters":[
        {
            "name":"input",
            "type":"List"
        }
    ],
    "Example":{
        "Scenario":"if you want to add 2 to 1.",
        "Parameters":{
        "input":[2,1]
        }
    }
}
/* Example 2 */
{
    "required_parameters": [
        {
            "name": "input",
            "type": "List"
        }
    ],
    "Example": {
        "Scenario": "if you want to subtract 2 from 1.",
        "Parameters": {
            "input": [1,2]
        }
    }
}
\end{lstlisting}
\caption{The tool instruction of FuncQA generated by \method.
  }
  \label{tab:example_funcqa}
\end{table*}\label{sec:appendix}
\end{document}